# Sensitivity analysis in decision circuits

**Debarun Bhattacharjya and Ross D. Shachter**
Department of Management Science and Engineering
Stanford University
Stanford, CA 94305, USA
E-mail: debarunb@stanford.edu, shachter@stanford.edu

## Abstract

Decision circuits have been developed to perform efficient evaluation of influence diagrams [Bhattacharjya and Shachter, 2007], building on the advances in arithmetic circuits for belief network inference [Darwiche, 2003]. In the process of model building and analysis, we perform sensitivity analysis to understand how the optimal solution changes in response to changes in the model. When sequential decision problems under uncertainty are represented as decision circuits, we can exploit the efficient solution process embodied in the decision circuit and the wealth of derivative information available to compute the value of information for the uncertainties in the problem and the effects of changes to model parameters on the value and the optimal strategy.

## 1 INTRODUCTION

Influence diagrams are powerful communication tools and computational aids for the analysis of practical decision problems [Howard and Matheson, 1984]. Decision circuits are a recent graphical representation that have been introduced for the efficient evaluation of influence diagrams [Bhattacharjya and Shachter, 2007]. In this paper, we show that they are also useful for efficient sensitivity analysis in influence diagrams.

The phrase *sensitivity analysis* refers, in general, to understanding how the output for a system varies as a result of changes in the system's input(s). For influence diagrams, one may be concerned about how the optimal solution and the certain equivalent ($CE$) change with respect to a change in the parameters, i.e. the probabilities and the utilities, or a change in the informational assumptions of the problem. There are many questions that we can use sensitivity analysis to answer. For example, suppose we vary one parameter keeping all other parameters constant. For what range of this varying parameter is the current optimal strategy still optimal? How does the value change as this parameter is varied? What if the structure of the influence diagram changes and an uncertainty that would not have been revealed before we make a decision is now observed before the first decision is made? These are only some of the queries on the model that we seek to answer.

Several issues in sensitivity analysis of Bayesian belief networks have been studied, using arithmetic circuits [Darwiche, 2003] for efficient solutions [Chan and Darwiche, 2002; 2004; 2006]. Our work builds on this body of research. Arithmetic circuits are graphical representations that have been shown to be efficient at performing inference on belief networks. Decision circuits promise similar benefits in the context of sequential decision problems.

In sections 2, 3 and 4 we briefly discuss some preliminaries, and review literature on circuits and sensitivity analysis in influence diagrams. In section 5, we introduce key ideas of performing sensitivity analysis with decision circuits for normal form influence diagrams, i.e. influence diagrams involving a single decision node with no parents. We show how to perform some basic sensitivity analysis with decision circuits, such as: plotting the certain equivalent in a one-way sensitivity analysis, finding the range of a parameter over which the current optimal stays optimal, and computing the value of information of uncertainties that are not affected by decisions, using partial derivatives. This serves as an introduction to section 6, where we present results for influence diagrams that may contain multiple decision nodes. We show the challenges and necessary modifications from the previous section. Finally, section 7 describes our conclusions and directions for future work.

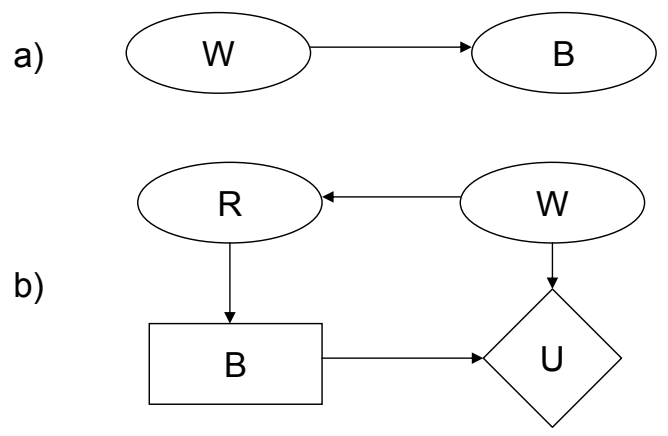

Figure 1: (a) A belief network; (b) An influence diagram.

## 2 PRELIMINARIES

We assume that the reader is familiar with graphical models such as Bayesian belief networks and influence diagrams (see Shachter (2007) for an overview). Consider the two examples shown in Figure 1. Figure 1a presents a belief network with two nodes, labelled W (Weather) and B (Bring umbrella). We are interested in knowing whether a friend will bring an umbrella, and we believe that it is easier to model this if we condition on the weather. Figure 1b shows the influence diagram for a decision problem in which our friend chooses whether to bring an umbrella based on her belief about the weather and her preferences, represented by the node U (Utility). She will observe a weather Report (R) before she makes her decision. We use these examples to demonstrate the concepts in later sections.

We make a distinction between *extensive* and *normal* form influence diagrams. When there is only one decision node and it has no parents, the diagram is said to be in normal form [Savage, 1954; Raiffa, 1968]. We extend that definition to allow for evidence, i.e. that we observe certain variables taking on specific values. Alternatively, when decisions are represented by separate nodes, or when the diagram has a decision node with at least one parent, the diagram is said to be in extensive form. There can be a large number of *strategies* in an influence diagram, one for each possible combination of observed uncertainties and decision alternatives. Although it is possible to convert any extensive form influence diagram into a normal form influence diagram, it may not be efficient to do so. The influence diagram shown in Figure 1b is in extensive form because the decision node B has a parent R.

In this paper we assume that the influence diagram has a single value node, referred to as utility node $U$. When making decisions, we choose the alternative that maximizes the probability that utility variable $U = 1$. The parents of utility node $U$, $\mathbf{pa(U)}$, are called the *value attributes* and we assess $P(U = 1|\mathbf{pa(U)}) = u(v(\mathbf{pa(U)}))$ where $u(.)$ is a von Neumann-Morgenstern utility function [von Neumann and Morgenstern, 1947] such that $U = 1$ is at least as good and $U = 0$ is at least as bad as anything that can happen, and $v(.)$ characterizes the value of the attributes in terms of a single numeraire, which we assume is dollars. Therefore, the *certain equivalent* ($CE$) of an uncertain $V$, given by $u^{-1}(E[u(V)])$, represents the certain payment that the decision-maker finds indifferent to $V$. Our sensitivity results will be expressed in terms of the certain equivalent because the utility values used for the internal computations have no intrinsic meaning. We also assume that the utility function $u(.)$ is strictly increasing and continuously differentiable. The most common utility functions are *linear*, $u(v) = av + b$, and *exponential*, $u(v) = -ae^{-v/\rho} + b$, where $a > 0$ and $\rho > 0$, both of which allow us to express the value of information exactly in closed form.

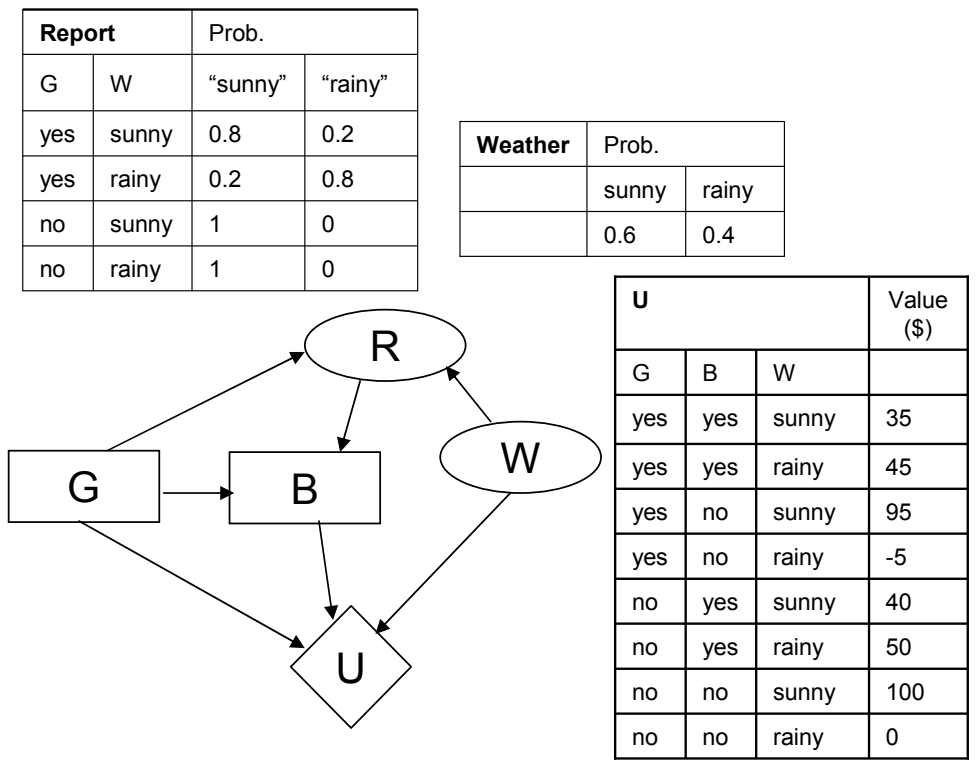

| **Report** | | Prob. | |
|---|---|---|---|
| G | W | "sunny" | "rainy" |
| yes | sunny | 0.8 | 0.2 |
| yes | rainy | 0.2 | 0.8 |
| no | sunny | 1 | 0 |
| no | rainy | 1 | 0 |

| **Weather** | Prob. | |
|---|---|---|
| | sunny | rainy |
| | 0.6 | 0.4 |

| **U** | | | Value ($) |
|---|---|---|---|
| G | B | W | |
| yes | yes | sunny | 35 |
| yes | yes | rainy | 45 |
| yes | no | sunny | 95 |
| yes | no | rainy | -5 |
| no | yes | sunny | 40 |
| no | yes | rainy | 50 |
| no | no | sunny | 100 |
| no | no | rainy | 0 |

Figure 2: An influence diagram example with numbers.

We will present some sensitivity analysis results for extensive form influence diagrams with the help of the influence diagram shown in Figure 2. In this example, our friend will decide whether to Gather evidence (G) and purchase a weather report. Her information gathering decision and the report will be known to her when she decides whether to bring her umbrella. If our friend does not gather evidence, the report is not informative and always states "sunny". The conditional probability tables and the value function $v(.)$ are shown in the figure. We assume that the decision maker has an exponential utility function $u(.)$ with risk tolerance $\rho = .02$. The optimal strategy is to gather evidence and bring the umbrella when the report says "rainy" and not to bring the umbrella when the report says "sunny". The $CE$ of the decision problem is \$52.5.

## 3 ARITHMETIC AND DECISION CIRCUITS

We review basic concepts regarding arithmetic and decision circuits in this section. Throughout this paper,

variables are denoted by upper-case letters ($X$) and their values by lower-case letters ($x$). A bold-faced letter indicates a set of variables. If $X$ is a variable with parents $\mathbf{Pa(X)}$, then $X\mathbf{Pa(X)}$ is called the family for variable $X$. The values of a binary variable $X$ are denoted $x$ and $\bar{x}$.

### 3.1 Arithmetic Circuits

Belief networks are associated with a unique multi-linear function over two kinds of variables, *evidence indicators* and *network parameters*. An evidence indicator $\lambda_x$ is a binary (0-1) variable, with $\lambda_x = 0$ whenever $X$ has been observed taking another value, i.e. it is observed not to be $x$. There is an evidence indicator associated with each possible instantiation $x$ of each network variable $X$. A network parameter $\theta_{x|\mathbf{pa(X)}}$ represents a conditional probability, $\theta_{x|\mathbf{pa(X)}} = P(x|\mathbf{pa(X)})$, and there is a network parameter for each possible instantation $x\mathbf{pa(X)}$ of family $X\mathbf{Pa(X)}$. Each *term* in the multi-linear function corresponds to one instantiation $\mathbf{z}$ of all the network variables $\mathbf{Z}$, involving the product of all evidence indicators and network parameters consistent with $\mathbf{z}$. The multi-linear function for a belief network is given as $f = \sum_{\mathbf{z}} \prod_{x\mathbf{pa(X)}\sim\mathbf{z}} \lambda_x \theta_{x|\mathbf{pa(X)}}$ where the sum is over every instantiation of all variables in the network and $x\mathbf{pa(X)} \sim \mathbf{z}$ represents all families consistent with $\mathbf{z}$. For example, consider the belief network of Figure 1a. Suppose that $W$ and $B$ are binary variables with states $w$ and $\bar{w}$, and $b$ and $\bar{b}$ respectively. The multi-linear function for this network is: $f = \lambda_w \lambda_b \theta_w \theta_{b|w} + \lambda_{\bar{w}} \lambda_b \theta_{\bar{w}} \theta_{b|\bar{w}} + \lambda_w \lambda_{\bar{b}} \theta_w \theta_{\bar{b}|w} + \lambda_{\bar{w}} \lambda_{\bar{b}} \theta_{\bar{w}} \theta_{\bar{b}|\bar{w}}$.

The multi-linear function is a useful construct for answering inference queries in belief networks. By setting the evidence indicators to 0 or 1, we can find the probability of observing any set of network variables $\mathbf{E}$. For instance, if we assign evidence to be $\mathbf{e} = \bar{b}$ by setting $\lambda_b = 0$ and all the other three evidence indicators as 1, the function returns $P(\bar{b}) = \theta_w \theta_{\bar{b}|w} + \theta_{\bar{w}} \theta_{\bar{b}|\bar{w}}$. In general, the evidence indicators help in summing the appropriate entries in the joint probability distribution table for computing the probability of the evidence, $P(\mathbf{e})$. Furthermore, the partial derivatives of the multi-linear function also provide solutions to several common probabilistic inference queries. We list two lemmas with important relationships between inference queries and the multi-linear function, as proven in Darwiche (2003):

**Lemma 1.** *For evidence* $\mathbf{e}$*, we have:* $P(\mathbf{e}) = f(\mathbf{e})$.

**Lemma 2.** *For every variable* $X$ *and evidence* $\mathbf{e}$ *such that* $X \notin \mathbf{E}$*, we have:* $P(x, \mathbf{e}) = \frac{\partial f}{\partial \lambda_x}(\mathbf{e})$.

Arithmetic circuits are graphical structures that efficiently represent, evaluate, and differentiate multi-linear functions. An arithmetic circuit is a rooted, directed acyclic graph whose leaf nodes are constants or variables and all other nodes represent either summation or multiplication. The *size* of an arithmetic circuit is the number of edges it contains.

The value of the multi-linear function is computed at the root of the circuit by *evaluating* the circuit in an *upward pass*, starting from the leaves and ending at the root. The result is denoted as $f(\mathbf{e})$, where $f(\mathbf{e}) = P(\mathbf{e})$ (see Lemma 1). We can calculate partial derivatives by *differentiating* the circuit through a subsequent *downward pass*, in which the parents are visited before the children. The upward and downward passes are also referred to as *sweeps*. For further details, please see Darwiche (2003).

Compact arithmetic circuits have been devised for belief networks that had previously been intractable [Darwiche, 2002; Chavira and Darwiche, 2005]. The circuit is compiled offline, where both the local structure (in the form of determinism and context-specific independence) as well as the conditional independencies of the graph at the global level are exploited. Several inference queries on the network can then be processed online, through subsequent operations on the compiled arithmetic circuit. Efficient sensitivity analysis of the belief network is possible with sweeps of the compiled circuit.

### 3.2 Decision Circuits

Decision circuits are arithmetic circuits augmented with maximization nodes. They represent the dynamic programming function corresponding to a sequential decision problem. The *size* of a decision circuit is the number of edges it contains. Figure 3 presents a decision circuit corresponding to the influence diagram shown in Figure 1b.

Decision circuits for influence diagrams can be constructed in the variable elimination order [Bhattacharjya and Shachter, 2007], similar to a construction technique for arithmetic circuits [Darwiche, 2000]. The evidence indicator $\lambda_d$ for decision $D$ is initialized to 0 only if the alternative $d$ is no longer available to the decision maker. The network parameter $\theta_{d|\mathbf{pa(D)}}$ for a decision $D$ is initialized to 1 if the alternative is conditionally available under scenario $\mathbf{pa(D)}$, and 0 otherwise. Once compiled, the decision circuit can be evaluated in an upward sweep analogous to evaluation in arithmetic circuits. Decision circuits are evaluated with evidence $\mathbf{e}' = \mathbf{e} \cup \{U = 1\}$ when $\mathbf{e}$ is observed. The best outcome $U = 1$ is also deemed to be observed since this is an *MEU* problem and therefore the goal is to find optimal policies that maximize the probability of the best outcome given the evidence. The value

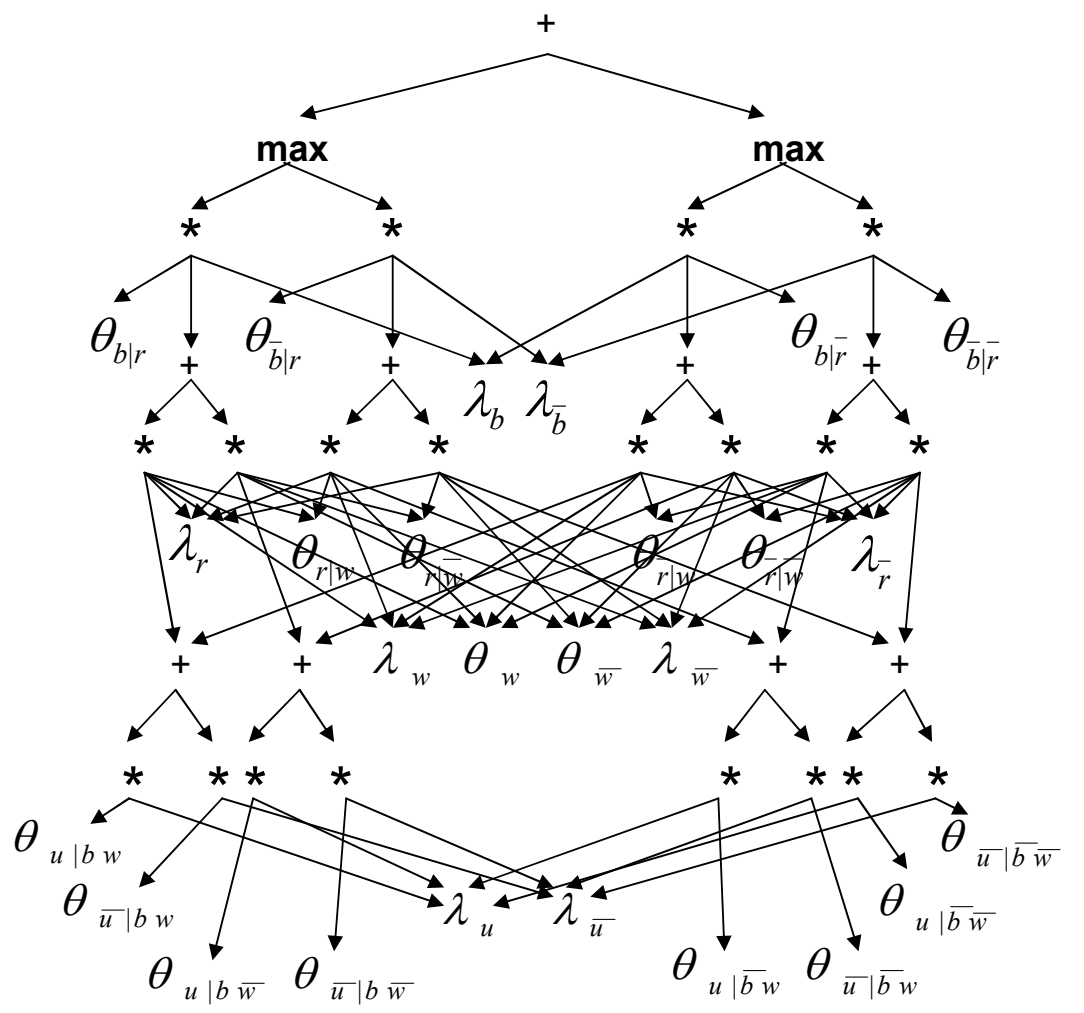

Figure 3: Decision circuit for the influence diagram shown in Figure 1b.

of the root node of the circuit is denoted $g(\mathbf{e}')$. The optimal strategy is computed on the upward sweep at the maximization nodes, where the alternative $d^*$ with the highest value is chosen, breaking ties arbitrarily. The network parameter $\theta_{d|\mathbf{u}}$ is set to 0 for all other alternatives $d$. The circuit can be differentiated in a subsequent downward sweep, by treating the maximization nodes as summation nodes. Although optimal policies are determined on the upward sweep, the $MEU$ and $CE$ are calculated also using results from the downward sweep. Specifically, from Bhattacharjya and Shachter (2007):

**Lemma 3.** $MEU = \frac{g(\mathbf{e}')}{\frac{\partial g}{\partial \lambda_u}(\mathbf{e}') + \frac{\partial g}{\partial \lambda_{\bar{u}}}(\mathbf{e}')}$.

**Lemma 4.** *For utility function* $u(.)$*, we have:* $CE = u^{-1}(MEU)$.

# 4 SENSITIVITY ANALYSIS IN INFLUENCE DIAGRAMS

The conditional probabilities in an influence diagram can be difficult to assess due to the paucity of data, expert judgments about key uncertainties, the decision maker's imprecision regarding preferences, and several other practical reasons. As a result, it is often beneficial to inspect the change in the outputs of the decision model based on variations in the inputs of the model. Such issues fall under the umbrella of sensitivity analysis.

Sensitivity analysis has been an essential aspect of decision analysis throughout the field's development. Sensitivity analysis aids in identifying the model's critical elements, forming the basis for iterative refinement of the model, and can also be used after the analysis for defending a particular strategy to the decision maker [Howard, 1983]. Sensitivity plots displaying the certain equivalents of different strategies can help identify the important variables. Sensitivity analysis for decision problems can be broadly classified in terms of whether one parameter is varied while others are kept constant (*one-way sensitivity analysis*) or when multiple parameters are simultaneously varied (*n-way sensitivity analysis*). One-way sensitivity analysis throws light upon the critical model variables whereas n-way sensitivity analysis provides insights into the general robustness of the model.

Sensitivity analysis in belief networks explores the sensitivity of inference queries such as the probability of evidence and conditional marginal probabilities given the evidence, to the conditional probabilities of the network [Laskey, 1995; Castillo et al, 1997; Kjaerulff and van der Gaag, 2000; van der Gaag and Renooij, 2001]. Sensitivity to inference queries using arithmetic circuits has also been studied [Chan and Darwiche, 2002; 2004] as has the sensitivity of Most Probable Explanations to parameter changes [Chan and Darwiche, 2006]. Our work differs from this line of research in that it uses decision circuits to determine sensitivity of the optimal strategy and the certain equivalent to parameter changes in decision problems.

To perform sensitivity analysis for decision problems, we will vary the model output with respect to *meta-parameters*, similar to Chan and Darwiche (2002) and Nielsen and Jensen (2003). For a variable $X$, we analyze sensitivity to all parameters of the form $\theta_{x|\mathbf{pa(X)}}$ as linear functions of a meta-parameter $\tau$. For example if $X$ is a binary variable with no parents, we can set $\theta_x = \tau$ and $\theta_{\bar{x}} = 1 - \tau$. In this paper we assume that there are $K$ meta-parameters, denoted $\tau_k, k = 1, ..., K$, each between 0 and 1, and that these meta-parameters are drawn explicitly in the decision circuit, although it is also possible to allow the meta-parameters to be implicit [Chan and Darwiche, 2002]. We assume that each meta-parameter is associated with only one variable. This ensures that the root value of the decision circuit is a piecewise-linear function of each meta-parameter. This can be extended to cases where the model inputs are non-linear functions of the meta-parameters, such as sensitivity to risk tolerance.

# 5 ANALYZING NORMAL FORM INFLUENCE DIAGRAMS

When there is only one decision node in an influence diagram and it has no parents, the diagram is said to be in normal form. We extend this definition to allow the observation that $\mathbf{E} = \mathbf{e}$. We discuss sensitivity analysis for normal form influence diagrams in this section.

### 5.1 Partial derivatives

Consider an influence diagram with a single decision node $D$ that has no parents. Let $X$ be an arbitrary uncertain variable, with parents $\mathbf{Pa}$ (Note that $D$ can be a parent of $X$). The dynamic programming function corresponding to this normal form influence diagram is given by $g = \max_d \theta_d \lambda_d \sum_{\mathbf{z}} \prod_{x,\mathbf{pa(X)} \sim \mathbf{z}} \theta_{x|\mathbf{pa(X)}} \lambda_x$ where $d$ is an alternative for decision node $D$. In normal form, each alternative represents a strategy. Let $EU_d$ be the expected utility for strategy $d$. Then $MEU = \max_d EU_d$, optimal strategy $d^* = argmax_d EU_d$ and $CE = u^{-1}(MEU)$. We will discuss many of the normal form sensitivity results using the following theorem, which outlines the significance of derivatives of $g(\mathbf{e}')$ and $g(\mathbf{e})$. These are the root values of the decision circuit evaluated at evidence $\mathbf{e}'$ and $\mathbf{e}$, respectively.

**Theorem 1.** *If evidence $\mathbf{e}' = \mathbf{e} \cup \{U = 1\}$ and $\mathbf{e}$ are swept through a decision circuit constructed for a normal form influence diagram, then for any node $v$:*
*(i)* $\frac{\partial EU_d}{\partial v} = \frac{\partial}{\partial v}\left[\frac{\frac{\partial g(\mathbf{e}')}{\partial \theta_d}}{g(\mathbf{e})}\right] = \frac{g(\mathbf{e})\frac{\partial^2 g(\mathbf{e}')}{\partial \theta_d \partial v} - \frac{\partial g(\mathbf{e})}{\partial v}\frac{\partial g(\mathbf{e}')}{\partial \theta_d}}{(g(\mathbf{e}))^2}$.
*(ii) If $MEU^{PI}$ is the maximal expected utility with perfect information on uncertainty X, then:* $MEU^{PI} = \sum_x \max_d \frac{\partial EU_d}{\partial \lambda_x}$.
*(iii)* $\frac{\partial CE_d}{\partial v} = \frac{\frac{\partial EU_d}{\partial v}}{u'(CE_d)}$.

*Proof.* (i): On the downward sweep the maximization nodes are treated as summation nodes, therefore $g = \sum_d \theta_d \lambda_d \sum_{\mathbf{z}} \prod_{x,\mathbf{pa(X)} \sim \mathbf{z}} \theta_{x|\mathbf{pa(X)}} \lambda_x$ for the downward sweep. Differentiating $g$ with respect to $\theta_d$ and evaluating at evidence $\mathbf{e}'$ results in $P(U = 1, \mathbf{e}|d)$ since only terms associated with alternative $d$ remain in the summation. The expected utility of alternative $d$ is $P(U = 1|\mathbf{e}, d)$, therefore $EU_d = \frac{1}{P(\mathbf{e})}\frac{\partial g(\mathbf{e}')}{\partial \theta_d}$. The result follows by recognizing that $P(\mathbf{e}) = g(\mathbf{e})$ ($\mathbf{e}$ is not responsive to any strategy $d$), and differentiating with respect to node $v$ using the quotient rule of differentiation.
(ii): $MEU^{PI} = \sum_x P(x|\mathbf{e}) \max_d P(U = 1|\mathbf{e}, x, d)$
$= \sum_x \max_d P(U = 1, x|\mathbf{e}, d)$
$= \sum_x \max_d \frac{\partial EU_d}{\partial \lambda_x}$.
We have used some results from the proof of part (i), and the fact that the uncertain variable $X$ is not affected by the decision, thus $P(x|\mathbf{e}) = P(x|\mathbf{e}, d) \forall d$.
(iii) $EU_d = u(CE_d)$; the result follows from differentiating both sides with respect to $v$ using the chain rule of differentiation. □

Theorem 1 presents a recipe for computing $\frac{\partial EU_d}{\partial v}$ and $\frac{\partial CE_d}{\partial v}$ for any node $v$ in the circuit, using $g(\mathbf{e})$, $g(\mathbf{e}')$ and their derivatives. The single and double derivatives of the form used in Theorem 1 can be obtained in time linear in the size of the circuit [Darwiche, 2000]. The double derivatives $\frac{\partial^2 g(\mathbf{e}')}{\partial \theta_d \partial v}$ are computed by sweeping $\frac{\partial g(\mathbf{e}')}{\partial \theta_d}$, for all strategies $d$, down the circuit. Thus the time complexity for obtaining $\frac{\partial EU_d}{\partial v}$ for all nodes is $O((N_S)(dc))$ where $N_S$ is the number of strategies and $dc$ is the size of the decision circuit. Note that the slope of $CE_d$ with respect to $v$ is not constant unless the utility function is linear; it will depend on the value of $CE_d$ in general.

### 5.2 One-way sensitivity analysis plots

In this sub-section, we show how to create a graph of the certain equivalent for the decision problem as a function of a particular meta-parameter when all others are kept at their reference values.

The expected utility for a particular strategy is a multi-linear function of all the meta-parameters. Therefore, the expected utility for a particular strategy $d$ is a linear function of a particular meta-parameter $\tau_k$, keeping all other meta-parameters at their reference values. We denote this linear function as $EU_d = \alpha_d(\tau_k)$. We already have a point on this line from the initial sweep used to evaluate the circuit, which we denote as $(\tau_k^0, \alpha_d^0)$. We also have the slope of this line, since it equals the partial derivative from part (i) of Theorem 1 choosing $v = \tau_k$. We denote the slope as $\alpha'_{d,k}$. Plotting the decision problem's $CE$ with respect to changes in $\tau_k$ entails applying the inverse utility function $u^{-1}(.)$ to the maximum of the lines for all strategies. If the required resolution is $\epsilon$, then the number of points required between 0 and 1 for the plot is $1/\epsilon$. The time complexity for preparing the plot for all meta-parameters, once we have the lines for expected utilities of all strategies with respect to all meta-parameters, is $O((\frac{1}{\epsilon})(K)(N_S))$ where $K$ is the number of meta-parameters and $N_S$ is the number of strategies.

Another approach to plotting the $CE$ against a meta-parameter is the standard method of *sample points*, where the decision problem is re-evaluated for every sample point over a range. If the required resolution is $\epsilon$, then the time complexity of this approach for obtaining plots for all meta-parameters is $O((\frac{1}{\epsilon})(K)(dc))$.

### 5.3 Admissible intervals

Another sensitivity question in the spirit of one-way sensitivity analysis is the following one. Suppose a certain meta-parameter is allowed to vary, keeping all other meta-parameters constant at their reference values. What is the range of this meta-parameter over which the current optimal strategy remains optimal? We call this range the *admissible interval* for a meta-parameter. In other words, we investigate how robust the optimal strategy is, with respect to changes in any meta-parameter.

The admissible intervals are easy to obtain for normal form influence diagrams, based on the ideas from the previous subsection. Since we have the lines for the expected utilities of all strategies with respect to all meta-parameters, we can obtain the admissible intervals $I_k$ for all meta-parameters simultaneously. Suppose $\Delta+$ and $\Delta-$ are the admissible positive and negative changes to $\tau_k$ from its reference value $\tau_k^0$ such that $d^*$ stays optimal. We present the following theorem, without proof, for computing $\Delta+$, $\Delta-$ and the intervals $I_k$. The proof entails finding the points at which another strategy overtakes $d^*$ and recognising that $\tau_k$ lies between 0 and 1. The time complexity for computing these intervals for all meta-parameters, once we have the lines for expected utilities of all strategies with respect to all meta-parameters, is $O((K)(N_S))$.

**Theorem 2.** *For meta-parameter $\tau_k$, the admissible positive change* $\Delta+ = \min_{ds.t.\alpha'_{d,k} > \alpha'_{d^*,k}} \left[\frac{\alpha^0_{d^*} - \alpha^0_d}{\alpha'_{d,k} - \alpha'_{d^*,k}}\right]$*, the admissible negative change* $\Delta- = \max_{ds.t.\alpha'_{d,k} < \alpha'_{d^*,k}} \left[\frac{\alpha^0_{d^*} - \alpha^0_d}{\alpha'_{d,k} - \alpha'_{d^*,k}}\right]$*, and the admissible interval* $I_k = \left[\max\left(0, \tau_k^0 + \Delta+\right), \min\left(1, \tau_k^0 + \Delta-\right)\right]$.

*Binary search* is another possible approach for finding the admissible intervals. The admissible interval for any meta-parameter $\tau_k$ is a convex set due to the linearity of the expected utilities for all strategies with respect to $\tau_k$. We can therefore locate the end-points of this interval by re-evaluating the circuit at points chosen by binary search to check if the current optimal strategy is still optimal. If the required resolution is $\epsilon$, then the number of sample points needed for the binary search is $O(-\ln(\epsilon))$. Thus the time complexity of obtaining admissible intervals for all meta-parameters by this method is $O((-\ln(\epsilon)(K)(dc))$.

### 5.4 Value of information

The value of information for a particular uncertainty is a useful sensitivity analysis query in sequential decision problems, specifying the maximum that the decision maker should be willing to pay to observe the uncertainty before making the first decision [Howard, 1966; Raiffa, 1968]. The double derivatives from part (i) of Theorem 1 can also compute the value of information for all uncertainties that are not affected by the decision, using part (ii) of Theorem 1 (choose $v = \lambda_x$). Here $MEU^{PI}$ is the maximal expected utility with perfect information on uncertain variable $X$. If the decision maker's utility function $u(.)$ is linear or exponential, then the value of information of the uncertain variable $X$ is the increase in the certain equivalent $u^{-1}(MEU^{PI}) - u^{-1}(MEU)$. In general, this is usually a good approximation for the value of information, even if the decision maker's utility function has a form other than linear or exponential [Raiffa, 1968].

Once we have the results from Theorem 1, computing the value of information involves summing and maximizing the partial derivatives. If the number of variables analyzed for value of information is $\sharp var$, and assuming that the maximum number of possibilities for all of these variables is bounded by some constant, then the time complexity for obtaining the value of information for all these variables is $O((N_S)(\sharp var))$.

## 6 ANALYZING EXTENSIVE FORM INFLUENCE DIAGRAMS

Sensitivity analysis in normal form influence diagrams is relatively easy to describe because we explicitly represent all the strategies. Here we discuss techniques for analyzing extensive form influence diagrams. All of these methods can also apply to normal form diagrams. We assume the standard conditions in the influence diagram literature such as "no forgetting" [Howard and Matheson, 1984; Shachter, 1986].

### 6.1 One-way sensitivity analysis plots

The sample points method from the previous section directly applies for extensive form influence diagrams as well. It is also possible to plot the $CE$ for a particular strategy $s$ against $\tau_k$ using decision circuits. To do this, the network parameters for decisions $\theta_{d|\mathbf{pa(D)}}$ need to be set according to $s$. Note that these have been preset according to the current optimal strategy $s^*$ after the initial upward sweep to evaluate the optimal strategy, hence they need to be reset if $s \neq s^*$. Once the network parameters for decisions are chosen appropriately, we can sweep up and down the decision circuit, treating the maximization nodes as summation nodes on both sweeps.

Figure 4 is a one-way sensitivity analysis plot for the influence diagram shown in Figure 2, with respect to meta-parameter $\tau_1$, the probability that the weather is sunny. The plot displays the variation in the $CE$ of the decision problem as well as the $CE$ of the current optimal strategy $s^*$ with respect to changes in $\tau_1$. We observe that for most values of the meta-parameter, the $CE$ for the decision problem does not vary too much, and thus the model is robust to small changes in $\tau_1$ around its current value of 0.6. The figure supports the intuition that for extreme values of the probability of sunshine, it is optimal to decide whether to bring

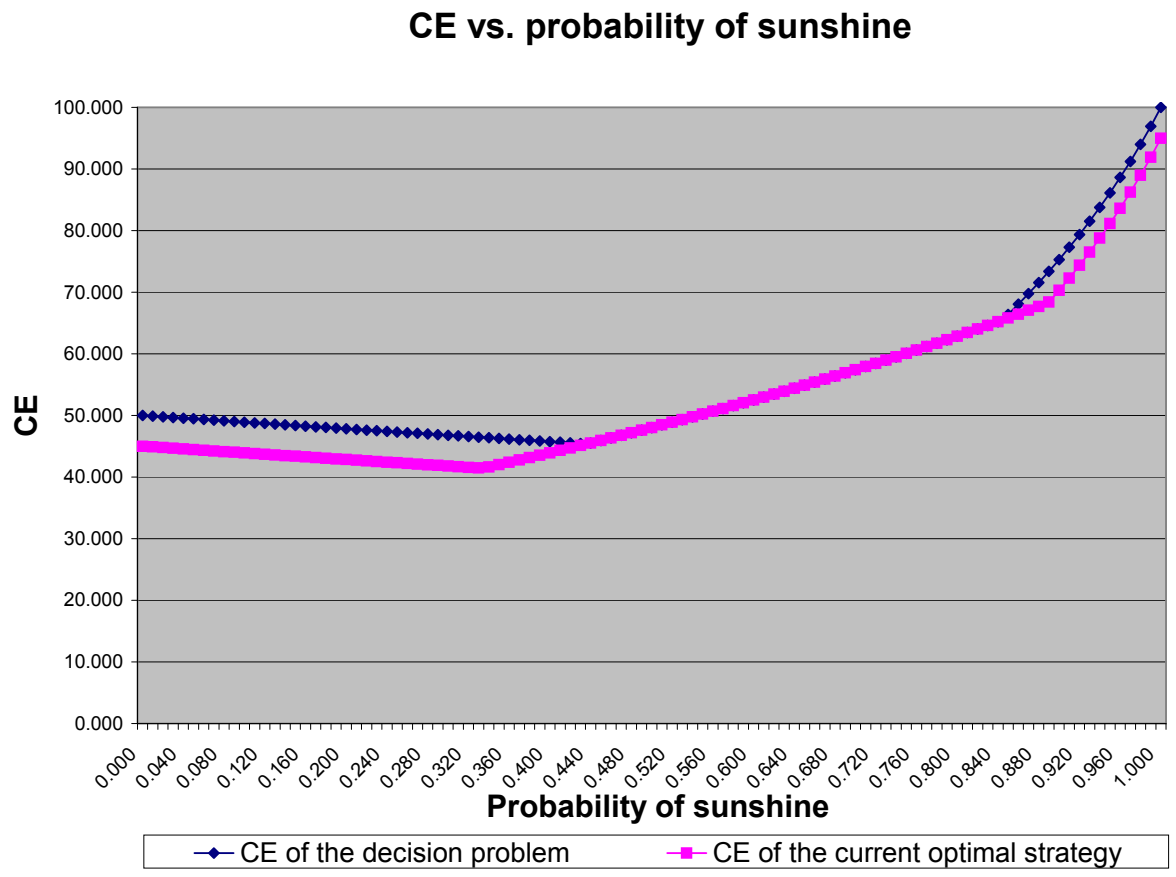

Figure 4: One way sensitivity analysis plot for an example.

the umbrella or not without paying for the evidence.

### 6.2 Admissible intervals

In this section, as an extension of our discussion in the section on normal form influence diagrams, we present the admissible interval algorithm, which returns bounds on the range over a particular meta-parameter for which the optimal strategy $s^*$ remains optimal when all other meta-parameters are kept at their reference values. The algorithm finds these bounds for all meta-parameters simultaneously. We assume that the optimal strategy $s^*$ and the $MEU$ have already been evaluated from initial sweeps through the decision circuit. The algorithm computes partial derivatives from a downward sweep starting from every maximization node. Before we describe the algorithm, we distinguish between *active* and *inactive max nodes* in the decision circuit. Inactive max nodes are those maximization nodes such that the derivative of the root with respect to these nodes, from the initial downward sweep, equals 0. This condition implies that an inactive node does not affect the expected utility of $s^*$, because there is 0 probability of being in that situation. Active max nodes are those max nodes in the circuit that are not inactive.

**Admissible Interval Algorithm:** Given a decision circuit constructed for an influence diagram and evaluated at evidence $\mathbf{e}'$, with optimal strategy $s^*$ and maximal expected utility $MEU$. Determine tight and weak bounds on admissible intervals, ${I_k}^T$ and ${I_k}^W$ for all meta-parameters $\tau_k$, $k = 1, 2, ..., K$.

1. Initialize all intervals ${I_k}^T$ and ${I_k}^W$ to $[0, 1]$.
2. Consider active max node $v$ in the circuit. Find the partial double derivatives for all alternatives (as described in section 5) by sweeping down the subcircuit rooted at $v$.
3. Find intervals for all meta-parameters over which the current optimal alternative remains optimal (as described in section 5). Take the intersection of these intervals with the corresponding intervals ${I_k}^W$ from the previous iteration and reset ${I_k}^W$ to these new intervals.
4. Repeat steps 2 and 3 above for inactive max node $v$ in the circuit, using ${I_k}^T$ in this case instead of ${I_k}^W$.
5. Repeat steps 2 and 3 above for all active and inactive max nodes in the decision circuit, updating the two corresponding kinds of intervals.
6. Set ${I_k}^T$ to be the intersection of the corresponding intervals ${I_k}^T$ and ${I_k}^W$ that were computed from previous steps.

If there are no inactive max nodes then clearly ${I_k}^W = {I_k}^T$, and this interval is the exact admissible interval. We now prove the correctness of this algorithm.

**Theorem 3.** *If $\tau_k$ lies in the interval ${I_k}^T$ then $s^*$ stays optimal and if $\tau_k$ does not lie in the interval ${I_k}^W$ then $s^*$ is no longer optimal.*

*Proof.* Consider any meta-parameter $\tau_k$ and its associated interval ${I_k}^T$. For $\tau_k \in {I_k}^T$, the equations in step 3 hold for all max nodes, since the resulting interval ${I_k}^T$ is the intersection of intervals. Thus $s^*$ is the optimal strategy when $\tau_k \in {I_k}^T$. Now if $\tau_k \notin {I_k}^W$, there must be an active max node $v$ such that the current optimal alternative for $v$ is no longer optimal. Therefore, $s^*$ can no longer be the optimal strategy. When $\tau_k \notin {I_k}^T$ and $\tau_k \in {I_k}^W$, we cannot be sure about whether $s^*$ stays optimal because it is possible for an inactive node to become active for the new value of $\tau_k$. □

Results from applying the algorithm to the example shown in Figure 2 are presented in Table 1. Two meta-parameters are considered for sensitivity analysis: the probability of the weather being sunny, $\tau_1 = \theta_w$, and the *specificity* of the report, $\tau_2 = \theta_{\bar{r}|\bar{w}}$, which is a measure of the expected number of false positives from the report. The table presents tight and weak intervals for both meta-parameters. The exact admissible interval for $\tau_1$ is $[0.44, 0.84]$, as can be seen from Figure 4. Note that this interval contains the tight interval but lies within the weak interval. The algorithm is able to compute the exact admissible interval for $\tau_2$ since the specificity of the report does not affect the value when evidence is not gathered.

Table 1: Results for the admissible interval algorithm for an example.

| **Metaparameter description** | **Tight interval** ($I_k{}^T$) | **Weak interval** ($I_k{}^W$) |
|---|---|---|
| $\tau_1 = \theta_w$ | $[.44, .67]$ | $[.44, .89]$ |
| $\tau_2 = \theta_{\bar{r}\mid\bar{w}}$ | $[.57, 1]$ | $[.57, 1]$ |

The admissible interval algorithm yields other useful results as a by-product. If the decision maker is interested in analyzing which alternative is optimal at a particular max node in the circuit, given that all other policies are made by the current optimal strategy, then this is easy to compute with the help of the intermediate steps in the algorithm. In other words, the algorithm helps in analyzing *neighbouring strategies* to $s^*$, defined as strategies that differ from $s^*$ only by an alternative in one active max node of the decision circuit. For instance, the decision maker could analyze the optimal policy at the first decision by comparing all strategies to $s^*$ that differ in only one alternative at a max node in the decision circuit corresponding to the first decision node in the influence diagram.

The binary search method from the previous section directly applies in extensive form influence diagrams, for finding admissible intervals for all meta-parameters. It computes these intervals exactly, one meta-parameter at a time. The admissible interval algorithm identifies bounds for these intervals, but for all meta-parameters simultaneously. We suggest that the analyst use the admissible interval algorithm as an initial test to identify potentially critical meta-parameters, and then use the binary search technique to focus attention on specific meta-parameters.

### 6.3 Value of information

The challenge of performing value of information analysis for extensive form influence diagrams is that it is difficult to keep track of all the derivatives for all the strategies. We propose a simple method to find the value of information for variables that are not affected by any decision, if they were to be observed before the first decision is made. The goal is to re-use the decision circuit that was formulated in the offline phase.

Assuming that the decision circuit has been evaluated and therefore that the probability of evidence $P(\mathbf{e})$ has already been computed, we can find the value of information for any variable $X$ with the help of some upward sweeps. If we pass the evidence $\mathbf{e}, U = 1, x$ in an upward sweep, once each for every possible value of $x$, we obtain $MEU^{PI}$ (the maximal expected utility with perfect information on uncertain variable $X$), as shown in the following theorem. We state the theorem without proof; the proof is similar to Theorem 1, part (ii).

**Theorem 4.** *If we sweep upward with evidence* $\mathbf{e}, U = 1, x$*, for every possible value of* $x$*, we obtain* $MEU^{PI} = \frac{1}{P(\mathbf{e})} \sum_x g(U = 1, x, \mathbf{e})$.

If the number of variables analyzed for value of information is $\sharp var$, and assuming that the maximum number of possibilities for all of these variables is bounded by some constant, then the time complexity for obtaining the value of information for all these variables is $O((dc)(\sharp var))$. For the example in Figure 2, the value of information on the weather is: $\$ \left[u^{-1}\left(g(U = 1, w) + g(U = 1, \bar{w})\right) - 52.5\right]$, which equals \$21.3. The report already provides some information about the weather, and further information is not worth more than \$21.3.

## 7 CONCLUSIONS AND FUTURE RESEARCH

Decision circuits are a graphical representation for the efficient analysis of influence diagrams, with the potential to exploit both the conditional independence in the graph as well as the local structure from asymmetry of real-world decision problems. Recent methods have been devised to create compact circuits using separability of the value function and operations such as pruning and coalescence [Bhattacharjya and Shachter, 2008]. If the analyst can compile an efficient decision circuit for an influence diagram, she can then use the compiled circuit to evaluate the optimal strategy and the $CE$, before performing sensitivity analyses of the kind demonstrated in this paper.

We explored several one-way sensitivity analysis queries and demonstrated techniques to answer them using decision circuits. We discussed sensitivity analysis for both normal form and extensive form influence diagrams. Sensitivity analysis is an essential technique to support decision problem modeling and it provides valuable insight about the critical assessments. $n$-way sensitivity analysis of influence diagrams can also be performed with decision circuits using higher order derivatives, similar to the techniques discussed in this paper and in Chan and Darwiche (2004). Likewise, the presented method for computing value of information can be extended to multiple variables, by sweeping evidence corresponding to every instantiation of the variables under consideration. Decision circuits allow a wide variety of queries about the model that can be addressed efficiently by decision circuit evaluation and differentiation.

**Acknowledgements**

This research was partially funded by the Energy Modeling Forum. We thank John Weyant for his support, Mark Chavira for helpful answers to our queries, Adnan Darwiche for providing relevant references and the anonymous reviewers for their feedback.

**References**

Bhattacharjya, D., and Shachter, R., 2007, Evaluating influence diagrams with decision circuits, In Parr, R., and van der Gaag, L., editors, *Proceedings of the Twenty-third Conference on Uncertainty in Artificial Intelligence*, pp. 9–16, Vancouver, BC, Canada: AUAI Press.

Bhattacharjya, D., and Shachter, R., 2008, Dynamic programming in decision circuits, Working paper.

Castillo, E., Gutierrez, J. M., and Hadi, A. S., 1997, Sensitivity analysis in discrete Bayesian networks, *IEEE Transactions on Systems, Man, and Cybernetics*, **27**, pp. 412-423.

Chan, H., and Darwiche, A., 2002, When do numbers really matter?, *Journal of Artificial Intelligence Research*, **17**, pp. 265–287.

Chan, H., and Darwiche, A., 2004, Sensitivity analysis in Bayesian networks: From single to multiple parameters, In Chickering, M., and Halpern, J., editors, *Proceedings of the Twentieth Conference on Uncertainty in Artificial Intelligence*, pp. 67–75, Banff, Canada: AUAI Press, Arlington, Virginia.

Chan, H., and Darwiche, A., 2006, On the robustness of Most Probable Explanations, In *Proceedings of the Twenty Second Conference on Uncertainty in Artificial Intelligence*, pp. 63–71, Cambridge, MA, USA: Morgan Kaufmann, San Mateo, California.

Chavira, M., and Darwiche, A., 2005, Compiling Bayesian networks with local structure, In *Proceedings of the Nineteenth International Joint Conference on Artificial Intelligence*, Edinburgh, Scotland, pp. 1306–1312.

Darwiche, A., 2000, A differential approach to inference in Bayesian networks, In Boutilier, C., and Goldszmidt, M., editors, *Proceedings of the Sixteenth Conference on Uncertainty in Artificial Intelligence*, pp. 123–132, Stanford, CA, USA: Morgan Kaufmann, San Mateo, California.

Darwiche, A., 2002, A logical approach to factoring belief networks, In *Proceedings of International Conference on Knowledge Representation and Reasoning*, pp. 409–420.

Darwiche, A., 2003, A differential approach to inference in Bayesian networks, *Journal of the ACM*, **50(3)**, pp. 280–305.

Howard, R., 1966, Information value theory, *IEEE transactions on Systems Science and Cybernetics*, **2**, pp. 22–26.

Howard, R., 1983, The evolution of decision analysis, In Howard, R., and Matheson, J., editors, 1989, *The Principles and Applications of Decision Analysis*, Strategic Decisions Group, Menlo Park, CA.

Howard, R., and Matheson, J., 1984, Influence diagrams, In Howard, R., and Matheson, J., editors, *Influence diagrams, belief nets, and decision analysis*, pp. 3–23, Wiley, Chichester.

Kjaerulff, U., van der Gaag, L. C., 2000, Making sensitivity analysis computationally efficient, In Boutilier, C., and Goldszmidt, M., editors, *Proceedings of the 16th Conference on Uncertainty in Artificial Intelligence*, pp. 317-325, Stanford, California: Morgan Kaufmann, San Francisco, California.

Laskey, K. B., 1995, Sensitivity analysis for probability assessments in Bayesian networks, *IEEE Transactions on Systems, Man, and Cybernetics*, **25**, pp. 901-909.

Nielsen, T., and Jensen, F. V., 2003, Sensitivity analysis in influence diagrams, *IEEE Transactions on Systems, Man, and Cybernetics - Part A: Systems and Humans*, **33** No. 1, March, pp. 223–234.

Raiffa, H., 1968, *Decision Analysis: Introductory Lectures on Choices under Uncertainty*, Addison-Wesley.

Savage, L., 1954, *The Foundations of Statistics*, Wiley, New York.

Shachter, R., 1986, Evaluating influence diagrams, *Operations Research*, **34** (November-December), pp. 871–882.

Shachter, R., 2007, Model building with belief networks and influence diagrams, In Edwards, W., Miles, R., and von Winterfeldt, D., editors, *Advances in Decision Analysis: From Foundations to Applications*, pp. 177-201, Cambridge University Press.

van der Gaag, L. C., and Renooij, S., 2001, Analysing sensitivity data from probabilistic networks, In Breese, J., and Koller, D., editors, *Proceedings of the 17th Conference on Uncertainty in Artificial Intelligence*, pp. 530-537, Seattle, WA: Morgan Kaufmann, San Francisco, California.

von Neumann, J., and Morgenstern, O., 1947, *Theory of Games and Economic Behavior*, 2nd edition, Princeton University Press, Princeton, NJ.